\documentclass{article}
\usepackage{spconf,amsmath,epsfig}
\usepackage{multicol}
\let\OLDthebibliography\thebibliography
\renewcommand\thebibliography[1]{
  \OLDthebibliography{#1}
  \setlength{\parskip}{0pt}
  \setlength{\itemsep}{0pt plus 0.3ex}
}

\begin{document}\sloppy

\def\x{{\mathbf x}}
\def\L{{\cal L}}

\title{A new way of video compression via forward-referencing using deep learning}
%
\name{Anonymous ICME submission}

\name{S.M.A.K. Rajin* , M. Murshed*, M. Paul**, S.W. Teng*, J. Ma*}
\address{*Federation University,\{s.rajin, manzur.murshed, shyh.wei.teng, j.ma\}@federation.edu.au 
**Charles Sturt University, mpaul@csu.edu.au}

\address{}

\maketitle

\begin{abstract}
To exploit high temporal correlations in video frames of the same scene, the current frame is predicted from the already-encoded reference frames using block-based motion estimation and compensation techniques. While this approach can efficiently exploit the translation motion of the moving objects, it is susceptible to other types of affine motion and object occlusion/deocclusion. Recently, deep learning has been used to model the high-level structure of human pose in specific actions from short videos and then generate virtual frames in future time by predicting the pose using a generative adversarial network (GAN). Therefore, modelling the high-level structure of human pose is able to exploit semantic correlation by predicting human actions and determining its trajectory. Video surveillance applications will benefit as stored “big” surveillance data can be compressed by estimating human pose trajectories and generating future frames through semantic correlation. This paper explores a new way of video coding by modelling human pose from the already-encoded frames and using the generated frame at the current time as an additional forward-referencing frame. It is expected that the proposed approach can overcome the limitations of the traditional backward-referencing frames by predicting the blocks containing the moving objects with lower residuals. Experimental results show that the proposed approach can achieve on average up to 2.83 dB PSNR gain and 25.93\% bitrate savings for high motion video sequences.
\end{abstract}

\section{Introduction}
\label{sec:intro}
Video compression is the process of encoding frames so that less space (bitrate) is used while still maintaining acceptable distortion (image quality). Video compression and transmission are considered fundamental functionalities of the multimedia driven Internet with applications in many fields such as video surveillance, social media, video streaming, and broadcasting. Traditional video compression has always used already-encoded data, that are also available to the decoder, to reconstruct the current frame \cite{richardson2011h}. To achieve this, each frame has to go through three main functional units:  prediction model, spatial model, and entropy encoder \cite{richardson2011h}.  The prediction model attempts to reduce redundancy in the current frame by exploiting the spatiotemporal correlations from the previously encoded frames (inter-prediction) or previously encoded neighboring pixels within the same frame (intra-prediction). We term this as \emph{backward-referencing} (BR) where the current frame is predicted from 'real' frames that have already been encoded i.e., looking backward w.r.t. the coding order of frames. The prediction model uses rate-distortion optimisation techniques to create a projection of the current frame with minimal bitrate for a fixed target image quality (variable bitrate (VBR) coder) or maximal image quality for a fixed target bitrate (constant bitrate (CBR) coder). 

While backward-referencing can efficiently exploit translation motion of the moving objects, it is vulnerable to other types of affine motion e.g., rotation, scaling, and shearing. It is also useless when parts of the moving objects are occluded (hidden in the reference frame) or deoccluded (reappear in the current frame). To address these limitations, researchers have considered multiple reference frames with some success e.g., the H.264/AVC video coding standard allows up to 16 reference frames to improve compression efficiency \cite{huang2006analysis}. Existing video coders are designed to compress data in a backward-referencing manner by making assumptions about next frames based on previously encoded ones. In this paper, we introduce a new way of video coding with \emph{forward-referencing} where next frames are reconstructed not only from previously encoded frames but also from 'virtual' frames that are artificially generated at current/future time by modelling the movement of moving objects using deep learning.

The recent success of deep learning in image recognition tasks has prompted research into their use within the field of video compression. Many companies have developed dedicated Neural Processing Units (NPUs) for mobile devices, which can process many trained Deep Neural Networks (DNNs) based applications in real-time. Even though training a DNN may suffer from a long delay, testing the network can be done in real-time \cite{tan2021efficient, niu2021grim}.  Various methods have been proposed in the literature that focused on leveraging neural network based coding models for image and video compression. Most of the changes or improvements to traditional video compression using a deep neural network are done either by improving certain modules of the existing codec \cite{chen2017deepcoder,liu2016cu,song2017neural} or by introducing pre- and post-processing \cite{lu2018deep} to the original and compressed frame sequence, respectively. A very few studies have tried to introduce pure learning-based video compression \cite{lu2019dvc}. Very recently, deep learning has been used to model the high-level structure of human pose in specific actions from short videos and then generate virtual frames in future time by predicting the pose using a generative adversarial network (GAN) \cite{villegas2017learning}.

Levering this recent advancement in virtual frame generation, we introduce a novel forward-referencing inter-prediction model, which complements the BR models used in the state-of-the-art video coding standards. The main contributions of our work in this paper are as follows: 
\begin{itemize}
  \item We propose a forward-referencing coding path using high-level structure (human pose) and deep visual-structure analogy as an additional prediction mode to the existing video coding standard. The proposed method uses an artificially generated future frame as an additional reference frame for inter-coding.
  \item Our novel architectural design compares and takes the best predictive model between the traditional standard and the proposed method for encoding the predictive (P) frames. If the proposed method produces better predictions, then it will be used for encoding; otherwise the traditional standards will be employed.
  \item For a long Group-of-Picture (GOP), which is highly desirable to exploit temporal redundancy maximally, the two anchor (I) frames differ greatly when objects are moving fast. For video sequences with fast-moving objects, we show that the proposed model outperforms the state-of-the-art video coding approach on several human action datasets.
\end{itemize}

\section{Proposed forward-referencing model}

\subsection{Forward referencing frame generation}
\begin{figure}[ht]
\begin{minipage}[b]{1.0\linewidth}
  \centering
 \centerline{\epsfig{figure=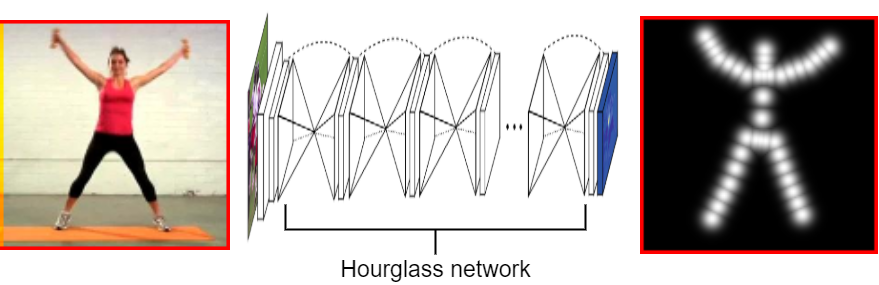,width=7.8cm}}
\end{minipage}
\caption{Human pose estimation on input image }
\label{fig:hourglass}
\end{figure}

The proposed forward-referencing frame generation model was developed based on \cite{villegas2017learning} hierarchical approach of generating future frames. The model consists of  two Neural Networks; the Hourglass Network \cite{newell2016stacked} and the Variational Auto Encoder Generative Adversarial Network (VAE-GAN) \cite{reed2015deep}. The Hourglass Network estimates the high-level structure of human pose in terms of 2D pose (x,y) coordinate position of 13 joints from RGB image as shown in Fig. \ref{fig:hourglass}. These poses and their corresponding images are then used to make the visual-structure analogy which follows as $p_{I\textendash frame}$ is to $p_t$ as $x_{I\textendash frame}$ is to $x_t$ , denoted as $p_{I\textendash frame} : p_t :: x_{I\textendash frame} : x_t$. Therefore, given that we have any of the three elements, reconstruction of the $4^{th}$ element is possible.

\begin{figure}[!ht]
\begin{minipage}[b]{1.0\linewidth}
  \centering
 \centerline{\epsfig{figure=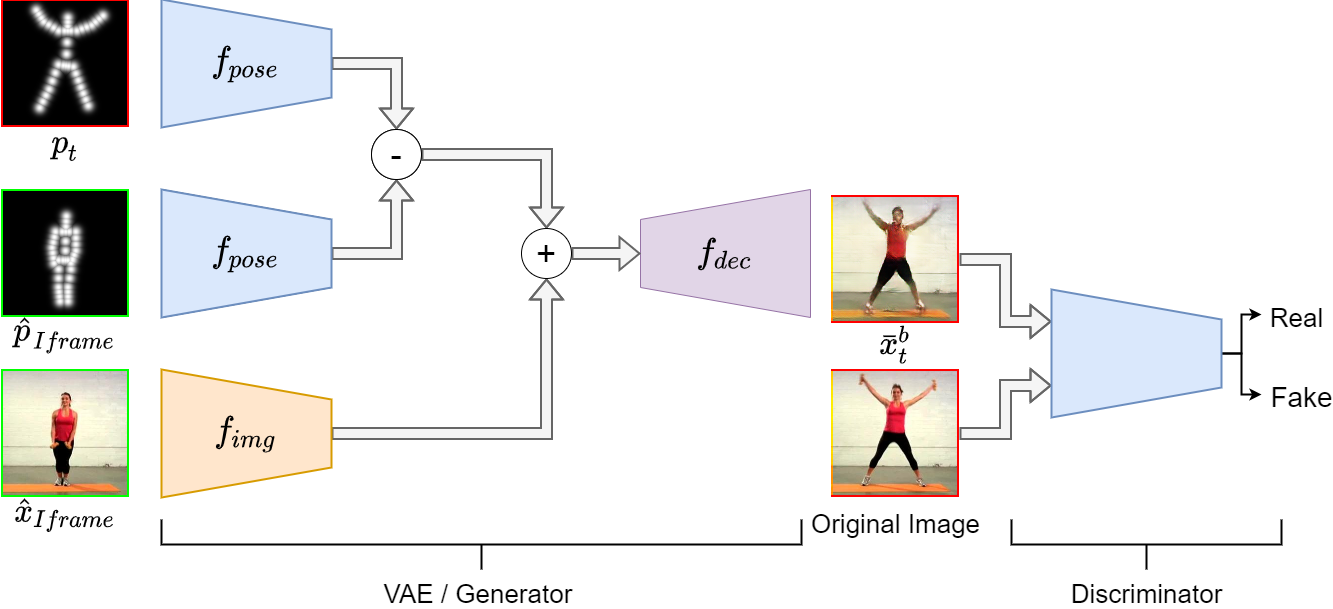,width=8.5cm}}
\end{minipage}
\caption{Illustration of the forward-referencing frame generator}
\label{fig:vaegan}
\end{figure}

The main success of our framework comes from the new idea of first making high-level structure estimation which allows us to send significant object information of the current frame to the decoder with minimal bits. The estimated pose structures of the current frame $p_t$, the I-frame $x_{I\textendash frame}$, and its corresponding estimated pose structure $p_{I\textendash frame}$ use visual structural analogy to synthesize the forward referencing frame. The visual-structure analogy can be performed by eq. \ref{eq1}
\begin{eqnarray}
\bar{x}_t = f_{dec} (f_{pose} ( g( p_{t} ) ) - (f_{pose}( g(p_{I\textendash frame}))  \nonumber\\+ f_{img}(x_{I\textendash frame}))
\label{eq1}
\end{eqnarray}

where $\bar{x}_t$ is the generated prediction image and $p_t$ is its corresponding pose, and, $x_{I\textendash frame}$ and $p_{I\textendash frame}$ are the reference image and its corresponding pose, respectively. The function $g(.)$ maps the output from the Hourglass Network into a depth-concatenated heatmap.  Fig. \ref{fig:vaegan} shows the network diagram of forward-referencing model.

In this paper, we only use the first frame (I-frame) for the reference image and pose. The pose coordinates are represented in 7 digits and to keep the consistency of the bit allocation for encoding poses, the highest possible bits require to represent all 26 xy-coordinates of a single frame can be calculated by 26 coordinate x 24 bits/ coordinate = 624 bits.

\subsection{Overview of the architecture}

\begin{figure}[ht]
\begin{minipage}[b]{1.0\linewidth}
  \centering
 \centerline{\epsfig{figure=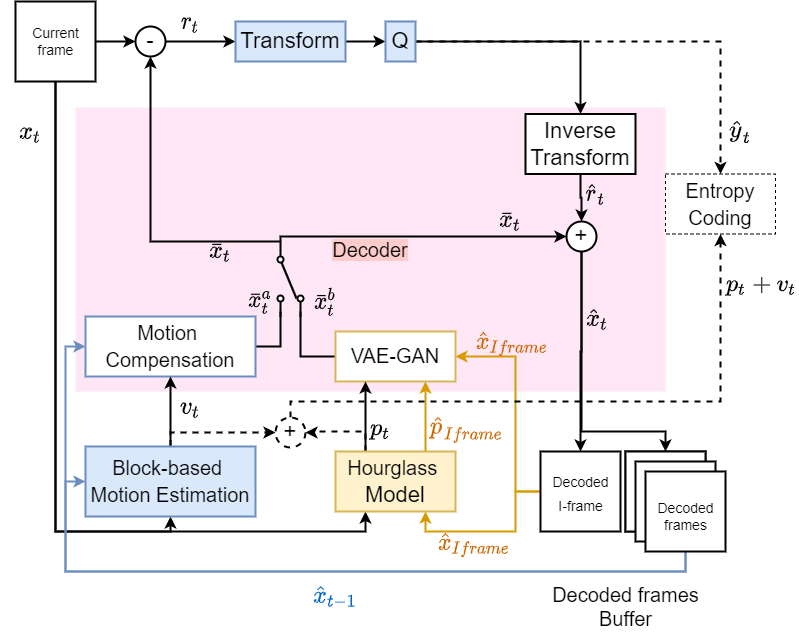,width=8.5cm}}
  
\end{minipage}
\caption{The proposed framework shows two additional components (in orange) to the traditional inter-predictive coding architecture}
\label{fig:Framework}
\end{figure}

Our model acts as an add-on to traditional inter-predictive coding architecture. The forward-referencing model was integrated into the traditional video coder to improve its coding efficiency. Our approach utilizes the two major components in addition to the traditional video encoder-decoder as shown in Fig. \ref{fig:Framework}. 

Let $V = \{x_1, x_2, x_3, ..., x_t, ...\}$ denote the current video sequence, where $x_t$ is the frame at time step $t$. The predicted frame is denoted as $\bar{x}_t$ and the decoded/reconstructed frame is denoted as $\hat{x}_t$. The residual frame, $r_t$ is the error (difference) between the original frame $x_t$ and the predicted frame $\bar{x}_t$. $\hat{r}_t$ represents the decoded/reconstructed residual frame.  Since the first frame in a Group Of Picture (GOP) is the I-frame, we can represent $x_1 = x_{I\textendash frame}$. Traditional approach of block-based motion compensated predicted frame is denoted as $\bar{x}_t^a$ and $v_t$ represents the motion vector The forward referencing frame is donated as $\bar{x}_t^b$ and $p_t$ is the estimated high-level structure (in our case 26 x y coordinates of the human pose) outputted by the Hourglass model for input images. To generate the future referencing frame, $\bar{x}_t^b$, inputs to the VAE-GAN are I-frame, $x_{I\textendash frame}$ and its' corresponding pose, $p_{I\textendash frame}$ and current frames' corresponding pose, $p_t$. The predicted frame, $\bar{x}_t$, is constructed using the rate-distortion optimization technique and the best macroblock is chosen between $\bar{x}_t^a$ and $\bar{x}_t^b$. $\hat{y}_t$ is the transformed and quantized version of $r_t$.

\section{Experimental results}
\subsection{Dataset for training and testing}
 We used the Penn Action dataset \cite{zhang2013actemes} to train, test and validate our model. The Penn Action dataset consists of 2326 video sequences of 15 different actions with 13 human pose joints annotation for each sequence. However, in total 1101 video sequences (611 for training and 490 for testing) of 8 different actions were used due to very noisy ground truth. These are baseball pitch (BP), baseball swing (BS), clean and jerk (CJ), golf swing (GF), jumping jacks (JJ), jump rope (JR), tennis forehand (TF), and tennis serve (TS). Frames are cropped based on temporal tubes to remove as much background as possible while making sure the foreground object i.e. the human of interest is in all frames. To train our VAE-GAN, we used the compound loss from \cite{dosovitskiy2016generating}, which can be defined by 
 \begin{eqnarray}
 L = L_{img} + L_{feat} + L_{Gen}
 \label{eq2}
 \end{eqnarray}

where $L_{img}$ is the loss in image space, $L_{feat}$ is the loss in feature space and. $L_{Gen}$ is the adversarial loss that allows our model to generate realistic-looking images. Each of these loss functions can be define as follows: 

\begin{eqnarray}
 L_{img} =  ||x_{t} - \hat{x}_{I\textendash frame} ||^2
\label{eq3}
\end{eqnarray}
\begin{eqnarray}
L_{feat} = ||  C_1 (x_{t}) - C_1 (\hat{x}_{I\textendash frame})||^2  \nonumber\\
           + ||  C_2 (x_{t}) - C_2 (\hat{x}_{I\textendash frame})||^2 
\label{eq4}
\end{eqnarray}
 \begin{eqnarray}
 L_{Gen} = -\log D([p_{t},\hat{x}_{I\textendash frame}])
\label{eq5}
\end{eqnarray}

where, $x_{t}$ and $\hat{x}_{t}$ are the target and predicted frames, respectively. $C_1 (.)$ and $C_2 (.)$ are functions that extracts features representing mostly image appearance and image structure, respectively. $p_{t}$ is the pose corresponding to $x_{t}$ and $D(.)$ is the discriminator network \cite{villegas2017learning}. During the optimization of $D(.)$, we use the mismatch term defined in \cite{reed2016generative}, which allows $D(.)$ to become sensitive to mismatch between the condition and the generation. The discriminator loss can be defined as 
\begin{eqnarray}
 L_{Gen} = -\log D([p_{t},{x}_{I\textendash frame}]) \nonumber\\
                -0.5\log (1-D([p_{t},\hat{x}_{I\textendash frame}])) \nonumber\\
                -0.5\log (1-D([p_{t},{x}_{t}]))
\label{eq6}
\end{eqnarray}

To illustrate our result, we chose five video sequences from the dataset corresponding to 3 different translation motion of moving object, which are fast moving (JJ \& TF), moderately moving (BP \& BS) and slow moving (GS) objects. We used H.264 video coding standard initially to compare the relative performance against our forward referencing model. Due to the nature of the coding. We denote the H.264 standard as the BR model. 

\begin{figure*}[!h]
\begin{minipage}[b]{1.0\linewidth}
  \centering
 \centerline{\epsfig{figure=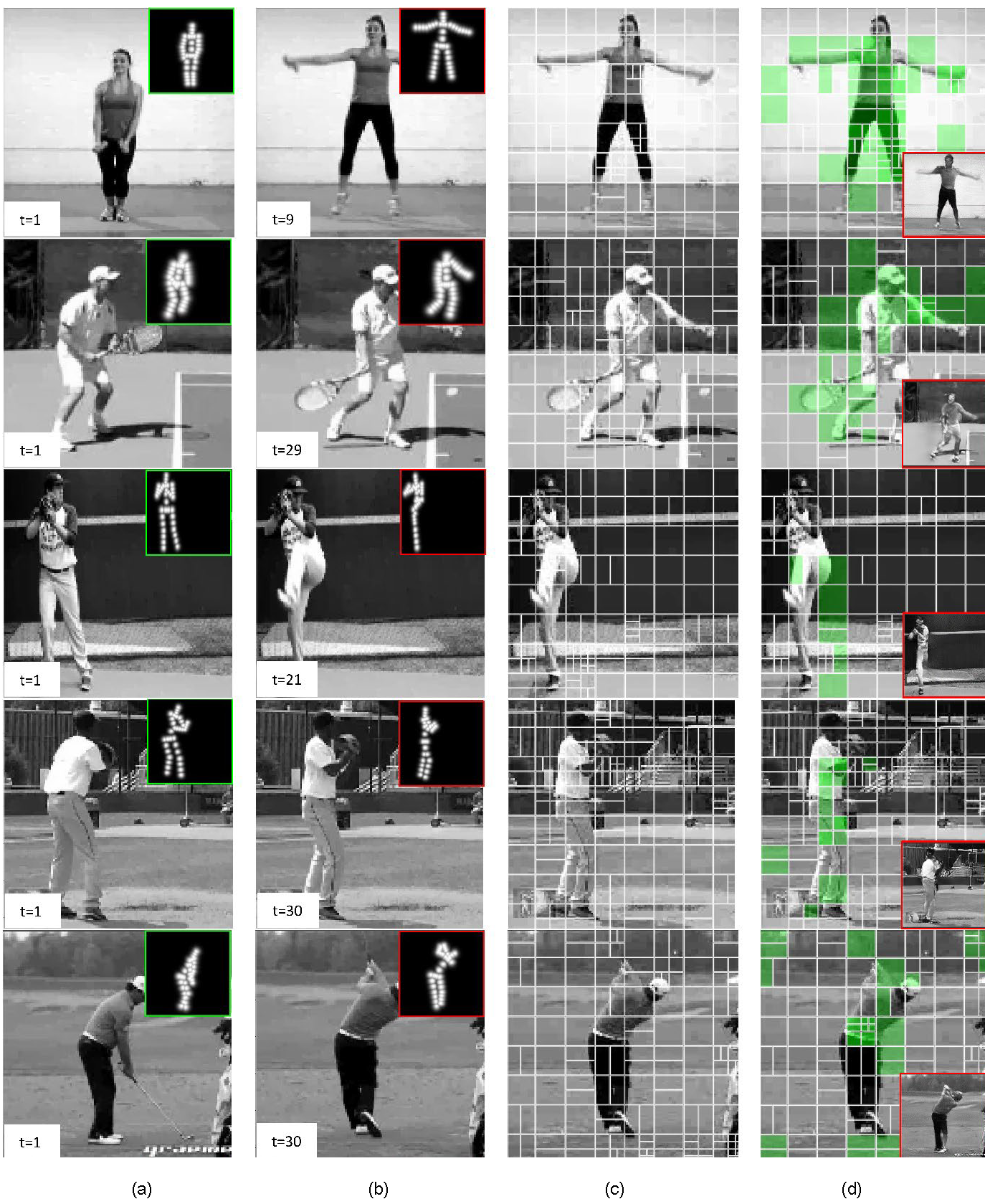,scale=.95}}

\end{minipage}
\caption{Visual illustration of how our model complements the BR model (a) I-frame is reconstructed using BR intra-coding. 
(b) is the original P-frame and its corresponding pose at the top-right corner (c) is the P-Frame encoded using BR and (d) is the P-Frame encoded using our forward referencing model. The highlighted area in green shows where our model outperforms the traditional inter-coding. GAN generated forward-referencing frame is shown at the bottom-right corner.}
\label{fig:frmodel}
\end{figure*}

\subsection{Comparison with traditional BR video coding}
\begin{figure}[ht]
\begin{minipage}[b]{1.0\linewidth}
  \centering
 \centerline{\epsfig{figure=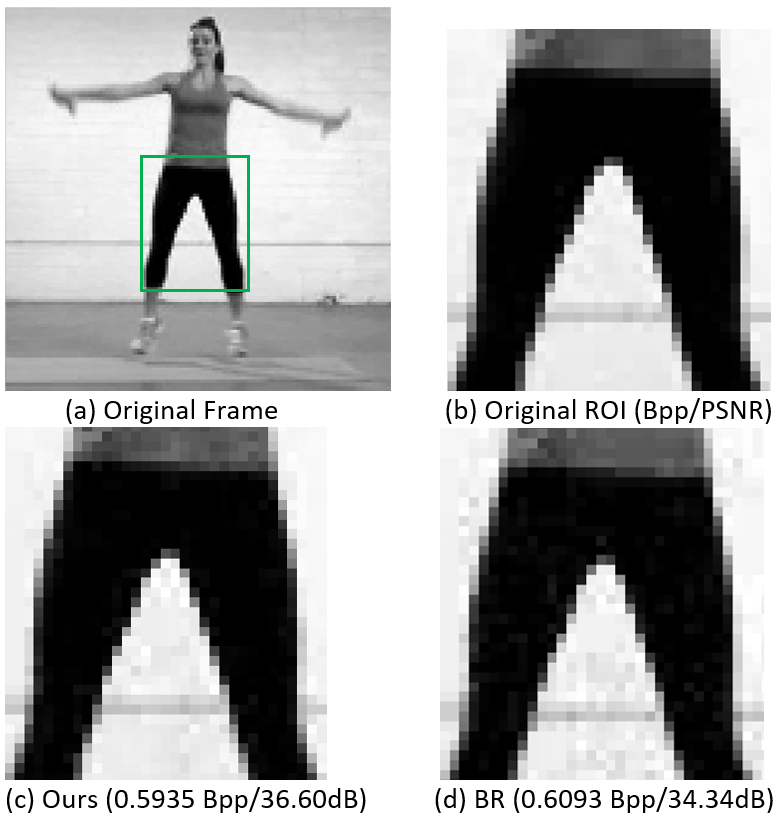, width=8.5cm}}

\end{minipage}
\caption{Visual quality of the reconstructed frames from different models. (a) is the original P-frame showing the region of interest (ROI) in green box. (b)-(d) are the original frame, reconstructed frame using our method and BR method.}
\label{fig:qality}
\end{figure}

 The first frame (I-frame) in a GOP is always intra-coded using the conventional approach and is the reference frame for the corresponding P-frame. To measure the effectiveness of our model, we encoded a sequence of P-frames, starting from frame t=2 up to t=32. Fig. \ref{fig:frmodel} shows the visual illustration of the mode selection process of the BR model compared to the mode selection when including our forward referencing model. The results shows the P-frames at t where the difference between the reference frame and its current frame is most significant. When the correlation between the two frames differ greatly, BR coding standards show some vulnerability, especially in the case of object occlusion/deocclusion. An example can be seen in Fig. \ref{fig:frmodel} (d), where the back of the golfer was not fully exposed in the reference I-frame. We calculate the sum of absolute difference (SAD) for each macroblock and sub macroblock of both reference frames, and the lowest SAD is used to reconstruct the current P frame. A subjective quality analysis on JJ can be seen in Fig. \ref{fig:qality}. Our proposed method (@QP=31) only consumes 0.5935 bpp while achieving the best perceptual quality which can be clearly observed between the legs where the texture is coarse for the BR model(@QP=34). We have kept the bitrate of BR method higher in order to take no undue advantages.
 
\begin{figure*}[!ht]
\begin{minipage}[b]{1.0\linewidth}
  \centering
\centerline{\epsfig{figure=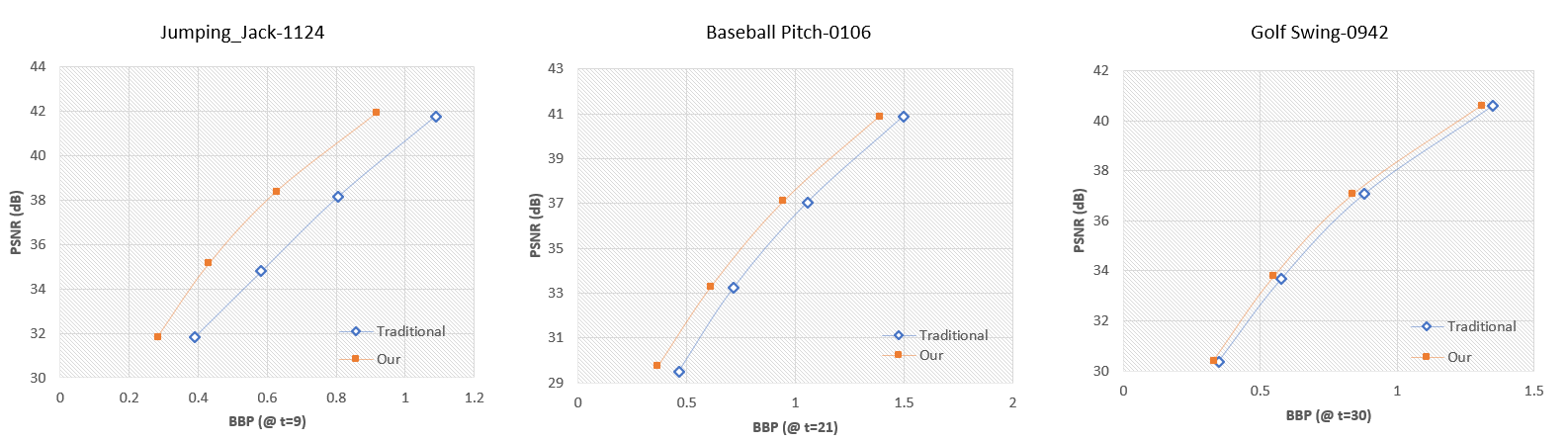,scale=.50}}
\end{minipage}

\caption{Rate-distortion (RD) performance comparison of the proposed technique with the traditional approach.}
\label{fig:RD}
\end{figure*}

Fig. \ref{fig:RD} shows the RD performance graph of three video sequence (JJ, BP,\& GF) using BR coding standards and our proposed method at QP = {24, 28, 34, and 38 }. Fig. \ref{fig:RD} reports the highest object motion difference between the I-frame and the predicted P-frame which occurs at times t=9, t=21, and t =30, respectively. Here the P-frame is encoded referencing both I-frame (traditional approach) and VAE-GAN generated frame. We calculate the average PSNR gains and bitrate saving using Bjontegaard metrics.

\begin{figure*}[!ht]
\begin{minipage}[b]{1.0\linewidth}
  \centering
 \centerline{\epsfig{figure=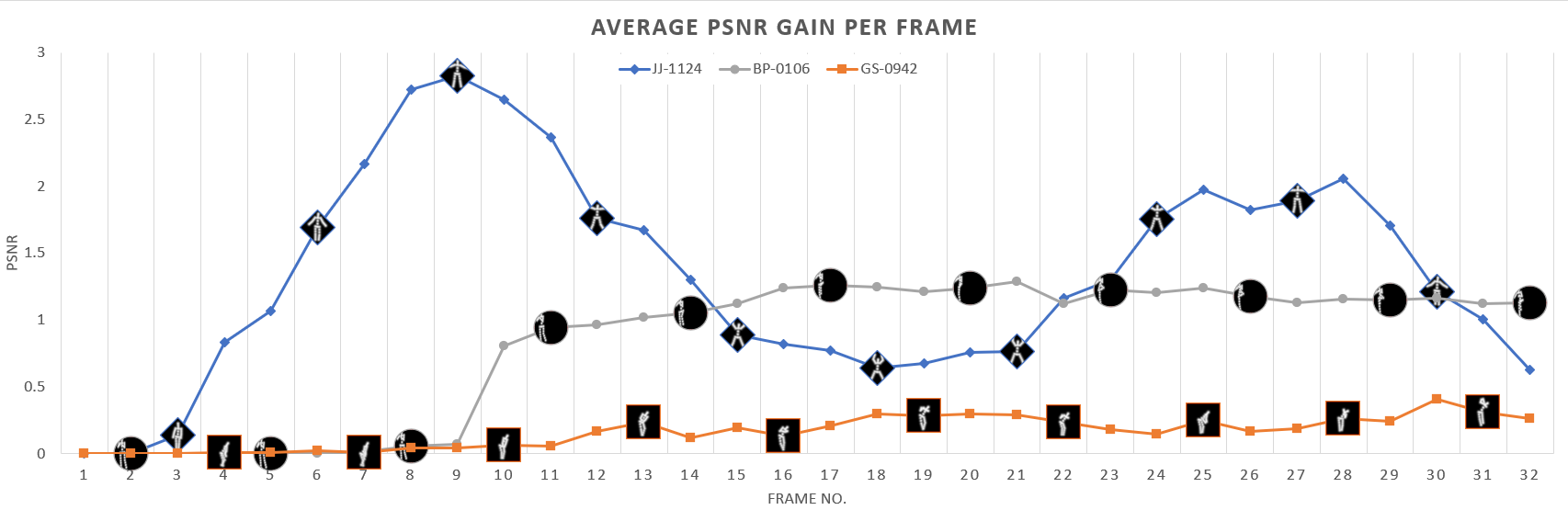, scale=.45}}
\end{minipage}
\caption{PSNR gain of each frame in GOP = 32}
\label{fig:psnrgain}
\end{figure*}

Fig. \ref{fig:psnrgain} shows the PSNR gain from our proposed model with respect to traditional video coding standards for 32 frames. It shows that the PSNR gain is dependent on the object movement and is at most when the pose of the object defers the most from the initial pose at t=1. It can be noted that among the three datasets, JJ-1124 has the highest PSNR gain of 2.83dB and bitrate saving of 25.93\% since the object in this video sequence has rapid movement.  Subsequently, GS-0942 has the lowest PSNR gain of 0.41 and bitrate saving of 5.26\% due to less motion of the object.
Table \ref{tab:t1} reflects the average PSNR gain and bitrate saving of the 32 frames for the three datasets compared to a traditional video coding standard. 

\begin{table}[ht]
\begin{center}
\caption{PSNR gain and bitrate saving for GOP=32} \label{tab:t1}
\begin{tabular}{|c|c|c|c|c|}
  \hline
  Dataset & Avg PSRN gain & Avg bitrate savings\\
  \hline
  JJ-1124 &	1.3887 & 14.36\%  \\
  BP-0106 & 0.8498 &  9.00\% \\
  GS-0942 & 0.1656 & 2.19\%	\\
  TS-2288 & 0.1646 & 2.13\% \\
  \hline
\end{tabular}
\end{center}
\end{table}

\section{Conclusions and Discussion}

In this paper, we present a new forward-referencing framework as an additional component to the SOTA video coding standards. Leveraging the advancement of deep generative models, we propose a new solution for video coding. In our experiment, we implement our framework on H.264 and compare the performance. One of the key features of our architecture is that the performance evaluation of encoding video sequences will not be less than the SOTA, rather in certain cases it will only improve the codec. As a future direction, we would like to explore more advanced deep neural network models and improve them to produce better quality forward-referencing frames. We will also implement our concept on HEVC and VCC codec in the future with higher resolution video sequences.

\bibliographystyle{IEEEbib}
\bibliography{icme2022template}

\begin{thebibliography}{10}

\bibitem{richardson2011h}
Iain~E Richardson,
\newblock {\em The H. 264 advanced video compression standard},
\newblock John Wiley \& Sons, 2011.

\bibitem{huang2006analysis}
Yu-Wen Huang, Bing-Yu Hsieh, Shao-Yi Chien, Shyh-Yih Ma, and Liang-Gee Chen,
\newblock ``Analysis and complexity reduction of multiple reference frames
  motion estimation in h. 264/avc,''
\newblock {\em IEEE Transactions on Circuits and Systems for Video Technology},
  vol. 16, no. 4, pp. 507--522, 2006.

\bibitem{tan2021efficient}
Tianxiang Tan and Guohong Cao,
\newblock ``Efficient execution of deep neural networks on mobile devices with
  npu,''
\newblock in {\em Proceedings of the 20th Int. Conf. on Information Processing
  in Sensor Networks (co-located with CPS-IoT Week 2021)}, 2021, pp. 283--298.

\bibitem{niu2021grim}
Wei Niu, Zhengang Li, Xiaolong Ma, Peiyan Dong, Gang Zhou, Xuehai Qian, Xue
  Lin, Yanzhi Wang, and Bin Ren,
\newblock ``Grim: A general, real-time deep learning inference framework for
  mobile devices based on fine-grained structured weight sparsity,''
\newblock {\em IEEE Transactions on Pattern Analysis and Machine Intelligence},
  2021.

\bibitem{chen2017deepcoder}
Tong Chen, Haojie Liu, Qiu Shen, Tao Yue, Xun Cao, and Zhan Ma,
\newblock ``Deepcoder: A deep neural network based video compression,''
\newblock in {\em 2017 IEEE Visual Communications and Image Processing (VCIP)}.
  IEEE, 2017, pp. 1--4.

\bibitem{liu2016cu}
Zhenyu Liu, Xianyu Yu, Yuan Gao, Shaolin Chen, Xiangyang Ji, and Dongsheng
  Wang,
\newblock ``Cu partition mode decision for hevc hardwired intra encoder using
  convolution neural network,''
\newblock {\em IEEE Transactions on Image Processing}, vol. 25, no. 11, pp.
  5088--5103, 2016.

\bibitem{song2017neural}
Rui Song, Dong Liu, Houqiang Li, and Feng Wu,
\newblock ``Neural network-based arithmetic coding of intra prediction modes in
  hevc,''
\newblock in {\em 2017 IEEE Visual Communications and Image Processing (VCIP)}.
  IEEE, 2017, pp. 1--4.

\bibitem{lu2018deep}
Guo Lu, Wanli Ouyang, Dong Xu, Xiaoyun Zhang, Zhiyong Gao, and Ming-Ting Sun,
\newblock ``Deep kalman filtering network for video compression artifact
  reduction,''
\newblock in {\em Proceedings of the European Conf. on Computer Vision (ECCV)},
  2018, pp. 568--584.

\bibitem{lu2019dvc}
Guo Lu, Wanli Ouyang, Dong Xu, Xiaoyun Zhang, Chunlei Cai, and Zhiyong Gao,
\newblock ``Dvc: An end-to-end deep video compression framework,''
\newblock in {\em Proceedings of the IEEE/CVF Conf. on Computer Vision and
  Pattern Recognition}, 2019, pp. 11006--11015.

\bibitem{villegas2017learning}
Ruben Villegas, Jimei Yang, Yuliang Zou, Sungryull Sohn, Xunyu Lin, and Honglak
  Lee,
\newblock ``Learning to generate long-term future via hierarchical
  prediction,''
\newblock in {\em Int. Conf. on machine learning}. PMLR, 2017, pp. 3560--3569.

\bibitem{newell2016stacked}
Alejandro Newell, Kaiyu Yang, and Jia Deng,
\newblock ``Stacked hourglass networks for human pose estimation,''
\newblock in {\em European Conf. on computer vision}. Springer, 2016, pp.
  483--499.

\bibitem{reed2015deep}
Scott~E Reed, Yi~Zhang, Yuting Zhang, and Honglak Lee,
\newblock ``Deep visual analogy-making,''
\newblock {\em Advances in neural information processing systems}, vol. 28, pp.
  1252--1260, 2015.

\bibitem{zhang2013actemes}
Weiyu Zhang, Menglong Zhu, and Konstantinos~G Derpanis,
\newblock ``From actemes to action: A strongly-supervised representation for
  detailed action understanding,''
\newblock in {\em Proceedings of the IEEE Int. Conf. on Computer Vision}, 2013,
  pp. 2248--2255.

\bibitem{dosovitskiy2016generating}
Alexey Dosovitskiy and Thomas Brox,
\newblock ``Generating images with perceptual similarity metrics based on deep
  networks,''
\newblock {\em Advances in neural information processing systems}, vol. 29,
  2016.

\bibitem{reed2016generative}
Scott Reed, Zeynep Akata, Xinchen Yan, Lajanugen Logeswaran, Bernt Schiele, and
  Honglak Lee,
\newblock ``Generative adversarial text to image synthesis,''
\newblock in {\em Int. Conf. on machine learning}. PMLR, 2016, pp. 1060--1069.

\end{thebibliography}

\end{document}